\documentclass[twoside]{article}

\usepackage[accepted]{aistats2022}
\usepackage[round]{natbib}

\usepackage[utf8]{inputenc} 
\usepackage[T1]{fontenc}    
\usepackage{color,xcolor}
\definecolor{mydarkblue}{rgb}{0,0.08,0.45}
\usepackage[colorlinks=true,
    linkcolor=mydarkblue,
    citecolor=mydarkblue,
    filecolor=mydarkblue,
    urlcolor=mydarkblue]{hyperref}
\usepackage{url}            
\usepackage{booktabs}       
\usepackage{nicefrac}       
\usepackage{microtype}      
\usepackage{amsmath,amsfonts,amssymb}
\usepackage{subfig}
\usepackage{mathtools}
\usepackage{enumitem}
\usepackage[page]{appendix}
\usepackage{algorithm}
\usepackage{algpseudocode}
\usepackage{float}

\usepackage{amsthm}
\usepackage{thmtools,thm-restate}
\newtheorem{remark}{Remark}
\DeclareMathOperator*{\argmin}{arg\,min}
\DeclareMathOperator{\Tr}{tr}
\newcommand*{\T}{\mathsf{T}}
\newcommand*{\W}{W}

\makeatother

\begin{document}

%

%

\twocolumn[

\aistatstitle{Towards Federated Bayesian Network Structure Learning with Continuous Optimization}

\aistatsauthor{Ignavier Ng \And Kun Zhang}

\aistatsaddress{ Carnegie Mellon University \And  Carnegie Mellon University\\Mohamed bin Zayed University of Artificial Intelligence} ]

\begin{abstract}
Traditionally, Bayesian network structure learning is often carried out at a central site, in which all data is gathered. However, in practice, data may be distributed across different parties (e.g., companies, devices) who intend to collectively learn a Bayesian network, but are not willing to disclose information related to their data owing to privacy or security concerns. In this work, we present a federated learning approach to estimate the structure of Bayesian network from data that is horizontally partitioned across different parties. We develop a distributed structure learning method based on continuous optimization, using the alternating direction method of multipliers (ADMM), such that only the model parameters have to be exchanged during the optimization process. We demonstrate the flexibility of our approach by adopting it for both linear and nonlinear cases. Experimental results on synthetic and real datasets show that it achieves an improved performance over the other methods, especially when there is a relatively large number of clients and each has a limited sample size.
\end{abstract}
\section{Introduction}\label{sec:intro}
Bayesian network structure learning (BNSL) is an important problem in machine learning and artificial intelligence \citep{Koller09probabilistic,Spirtes2001causation,Pearl2009causality,Peters2017elements}, and has been widely adopted in different areas such as healthcare \citep{Lucas2004bayesian} and Earth system science \citep{Runge2019inferring}. Traditionally, BNSL is often carried out at a central site, in which all data is gathered. With the rapid development of technology and internet, it has become increasingly easy to collect data. Therefore, in practice, data is usually owned by and distributed across a number of parties, which range from mobile devices and individuals to companies and hospitals.

In many cases, these parties, which are referred to as \emph{clients}, may not have a sufficient number of samples to learn a meaningful Bayesian network (BN) on their own. They may intend to collectively obtain aggregated  knowledge about the BN structure, but are not willing to disclose information related to their data owing to privacy or security concerns. Consider an example in which a number of hospitals wish to collaborate in learning a BN to discover the conditional independence structure underlying their variables of interests, e.g., medical conditions. Clearly, they are not allowed to share their patients' records because of privacy regulations. Therefore, the sensitivity nature of the data has made it infeasible to gather the data from different clients to a central site for BNSL.

Several approaches have been developed to learn BN structures in a distributed fashion. A natural approach is that each client estimates its BN structure independently using its local dataset, and share it to a central server, which applies some heuristics, e.g., voting \citep{Na2010distributed}, to aggregate these estimated structures. However, this approach may lead to suboptimal performance as the information exchange among clients is rather limited, and the estimated BNs by the individual clients may not be accurate. It remains a challenge to develop an approach that integrates the local information from different clients for learning a global BN structure while not exposing the clients' local datasets.

A principled federated learning strategy is then needed. In the past few years, \emph{federated learning} has received attention in which many clients collectively train a machine learning model based on a certain coordination strategy. Each client does not have to exchange its raw data; instead, they disclose only the minimal information necessary for the specific learning task, e.g., model parameters and gradient updates, which have been shown to work well in many tasks, e.g., image classification \citep{McMahan2017communication} and recommender system \citep{Chai2020secure}. We refer the reader to \citep{Yang2019federated,Li2020federated,Kairouz2021advances} for further details and a review on federated learning.

Most of the recent federated learning approaches, e.g., federated averaging \citep{McMahan2017communication}, are based on continuous optimization, which may not straightforwardly apply to standard score-based BNSL methods that rely on discrete optimization, e.g., dynamic programming \citep{Singh2005finding}, greedy search \citep{Chickering2002optimal}, and integer programming \citep{Cussens2011bayesian}. On the other hand, continuous optimization methods for BNSL have been recently developed \citep{Zheng2018notears,Zheng2020learning} by utilizing an algebraic characterization of acyclicity, which provides an opportunity to render federated learning possible for BNSL.

{\bf Contributions.} \ \ \ 
In this work, we present a federated learning approach to estimate the structure of BN from data that is horizontally partitioned across different parties. Our contributions are as follows:
\vspace{-0.1em}
\begin{itemize}
    \setlength\itemsep{0.0em}
    \item We propose a distributed BNSL method based on continuous optimization, using the alternating direction method of multipliers (ADMM), such that only the model parameters have to be exchanged during the optimization process.
    \item We demonstrate the flexibility of our approach by adopting it for both linear and nonlinear cases.
    \item We conduct experiments to validate the effectiveness of our approach on synthetic and real datasets.
\end{itemize}
\vspace{-0.1em}

{\bf Organization of the paper.} \ \ \ 
We describe the related work in Section \ref{sec:related_work}, and the problem formulation in Section \ref{sec:problem_formulation}. In Section \ref{sec:suff_stats_encryption}, we describe a simple privacy-preserving approach to learn linear Gaussian BNs and its possible drawbacks. We then present the proposed federated BNSL approach in Section \ref{sec:federated_bnsl}. We provide empirical results to validate our approach in Section \ref{sec:exp}, and a discussion in Section \ref{sec:discussion}.
\section{Related Work}\label{sec:related_work}
We provide a review on various aspects of BNSL that are relevant to our problem formulation and approach.

\subsection{BNSL with Continuous Optimization}\label{sec:related_work_continuous_optimization}
Recently, \citet{Zheng2018notears} developed a continuous optimization method to estimate the structure of linear BNs subject to an equality acyclicity constraint. This method has been extended to handle nonlinear models \citep{Kalainathan2018sam, Yu19daggnn, Ng2019graph, Ng2022masked, Lachapelle2020grandag, Zheng2020learning}, interventional data \citep{brouillard2020differentiable}, confounders \citep{Bhattacharya2020differentiable}, and time series \citep{Pamfil2020dynotears}. Using likelihood-based objective, \citet{Ng2020role} formulated the problem as an unconstrained optimization problem involving only soft constraints, while \citet{Yu2021nocurl} developed an equivalent representation of directed acyclic graphs (DAGs) that allows continuous optimization in the DAG space without the need of an equality constraint. In this work, we adopt the methods proposed by \citet{Zheng2018notears,Zheng2020learning} due to its popularity, although it is also possible to incorporate the other methods.

\subsection{BNSL from Overlapping Variables}\label{sec:related_work_overlapping_variables}
Another line of work aims to estimate the structures over the integrated set of variables from several datasets, each of which contains only samples for a subset of variables. \citet{Danks2008integrating} proposed a method that first independently estimates a partial ancestral graph (PAG) from each individual dataset, and then finds the PAGs over the complete set of variables which share the same d-connection and d-separation relations with the estimated individual PAGs. \citet{Triantafillou2015constraint,Tillman2014learning,Triantafillou2010learning} adopted a similar procedure, except that they convert the constraints to SAT solvers to improve scalability. The estimated graphs by these methods are not unique and often have a large indeterminacy. To avoid such indeterminacy, \citet{Huang2020causal} developed a method based on the  linear non-Gaussian model \citep{Shimizu2006lingam}, which is able to uniquely identify the DAG structure. These methods mostly consider non-identical variable sets, whereas in this work, we consider the setup in which the clients have a set of identical variables, and hope to collective learn the BN structure in a privacy preserving way.

\subsection{Privacy-Preserving \& Distributed BNSL}\label{sec:related_work_distributed_bnsl}
To learn BN structures from horizontally partitioned data, \citet{Gou2007learning} adopted a two-step procedure that first estimates the BN structures independently using each client's local dataset, and then applies further conditional independence test. Instead of using statistical test in the second step, \citet{Na2010distributed} used a voting scheme to pick those edges identified by more than half of the clients. These methods leverage only the final graphs independently estimated from each local dataset, which may lead to suboptimal performance as the information exchange may be rather limited. Furthermore, \citet{Samet2009privacy} developed a privacy-preserving method based on secure multiparty computation, but is limited to the discrete case.

For vertically partitioned data, \citet{Wright2004privacy,Yang2006privacy} constructed an approximation to the score function in the discrete case and adopted secure multiparty computation. \citet{Chen2003learning} developed a four-step procedure that involves transmitting a subset of samples from each client to a central site, which may lead to privacy concern.
\section{Problem Formulation}\label{sec:problem_formulation}
Let $G=(V,E)$ be a DAG that represents the structure of a BN defined over the random vector $X=(X_1,\dots,X_d)$ with a probability distribution $P(X)$. The vertex set $V$ corresponds to the set of random variables $\{X_1,\dots,X_d\}$ and the edge set $E$ represents the directed edges in the DAG $G$. The distribution $P(X)$ satisfies the Markov assumption w.r.t. the DAG $G$. In this paper, we focus on the \emph{linear BN} given by $X=B^\T X+N$, where $B$ is a weighted adjacency matrix whose nonzero coefficients correspond to the directed edges in $G$, and $N$ is a random noise vector whose entries are mutually independent. We use the term \emph{linear Gaussian BN} to refer to the linear BN in which the entries of the noise vector $N$ are Gaussian.

We consider the setting with a fixed set of $K$ clients in total, each of which owns its local dataset. The $k$-th client holds $n_k$ i.i.d. samples from the distribution $P(X)$, denoted as $\mathbf{x}_k=\{x_{k,i}\}_{i=1}^{n_k}$. Let $n=\sum_{k=1}^{K}n_k$ be the total sample size given by the sum of the sub-sample sizes $n_k,k=1,\dots,K$. We assume here that the data from different clients follows the same distribution, and leave the non-i.i.d. setting, which may have much potential, for future investigation. Given the collection of datasets $\mathbf{x}=\bigcup_{k=1}^{K}\mathbf{x}_k$ from different clients, the goal is to recover the true DAG $G$ in a privacy-preserving way. We focus on the setup in which the clients have incentives to collaborate in learning the BN structure, but are only willing to disclose minimal information (e.g., model parameters and estimated DAGs) related to their local datasets. The entire dataset $\mathbf{x}$ is said to be \emph{horizontally partitioned} across different clients, which implies that each local dataset shares the same set of variables but varies in samples. We also assume the existence of a central server that coordinates the learning process and adheres to the protocol.

\section{Computing Sufficient Statistics with Secure Computation}\label{sec:suff_stats_encryption}
Apart from the methods reviewed in Section \ref{sec:related_work_distributed_bnsl}, in this section we describe a simple approach based on secure computation to learn the structure of linear Gaussian BNs, and discuss its possible drawbacks.

{\bf Sufficient statistics.} \ \ \ 
Denote the empirical mean and covariance of $\mathbf{x}$ by $\mu_\mathbf{x}=(1/n)\sum_{k=1}^{K}\sum_{i=1}^{n_k}x_{k,i}$ and $\Sigma_\mathbf{x}=(1/n)\sum_{k=1}^{K}\sum_{i=1}^{n_k}x_{k,i}x_{k,i}^\T-\mu_\mathbf{x}\mu_\mathbf{x}^\T$, respectively. First notice that the empirical covariance matrix $\Sigma_\mathbf{x}$ and the total sample size $n$ are sufficient statistics for learning linear Gaussian BNs via constraint-based and score-based methods, formally stated as follows.
\begin{remark}\label{remark:suff_stats}
Suppose that the ground truth DAG $G$ and the distribution $P(X)$ form a linear Gaussian BN. Given $n$ samples $\mathbf{x}$ from the distribution $P(X)$, the empirical covariance matrix $\Sigma_\mathbf{x}$ and the sample size $n$ are sufficient statistics for constraint-based methods that rely on partial correlation tests and score-based methods that rely on the BIC score.
\end{remark}
This is because the first term of the BIC score \citep{Schwarz1978estimating} of linear Gaussian BNs corresponds to the maximum likelihood of multivariate Gaussian distribution with zero mean, for which the empirical covariance matrix is a sufficient statistic, and its second term involves the (logarithm of) sample size as the coefficient of model complexity penalty. Similarly, partial correlation tests are completely determined by these statistics. This implies that with the empirical covariance matrix $\Sigma_\mathbf{x}$ and sample size $n$, the samples $\mathbf{x}$ contain no additional information for constraint-based (e.g., PC \citep{Spirtes1991pc}) and score-based (e.g., GES \citep{Chickering2002optimal}, dynamic programming \citep{Singh2005finding}) methods, as well as for some continuous optimization based methods (e.g., NOTEARS \citep{Zheng2018notears}, GOLEM \citep{Ng2020role}).

Notice that the first and second terms of the empirical covariance matrix $\Sigma_\mathbf{x}$ are decomposable w.r.t. different samples. Therefore, each client can compute the statistics $\sum_{i=1}^{n_k}x_{k,i}$ and $\sum_{i=1}^{n_k}x_{k,i}x_{k,i}^\T$ of its local dataset $\mathbf{x}_k$, and send them, along with the sample size $n_k$, to the central server. The server then aggregates these values to compute the sufficient statistics $\Sigma_\mathbf{x}$ and $n$, which could be used to learn linear Gaussian BN via constraint-based or score-based methods.

{\bf Secure computation.} \ \ \ 
Each client must not directly share the statistics of its local dataset, since it may give rise to privacy issue. A privacy-preserving sharing approach is to adopt \emph{secure multiparty computation}, which allows different clients to collectively compute a function over their inputs while keeping them private. In particular, \emph{secure multiparty addition} protocols can be employed here to compute the aggregated sums of these statistics $\sum_{i=1}^{n_k}x_{k,i}$, $\sum_{i=1}^{n_k}x_{k,i}x_{k,i}^\T$, and $n_k$ from different clients, for which a large number of methods have been developed, including homomorphic encryption \citep{Paillier1999public,Gergely2011diffeRentially,Hazay2017efficient,Shi2011privacy}, secret sharing \citep{Shamir1979share,Burkhart2010sepia}, and perturbation-based methods \citep{Clifton2002tools}; we omit the details here and refer the interested reader to \citep{Goryczka2017comprehensive} for further details and a comparison. The aggregated statistics could then be used to compute $\Sigma_\mathbf{x}$ and $n$, and estimate the structure of linear Gaussian BN. 

While the approach of every client collectively computing the sufficient statistics with secure multiparty computation may seem appealing, a possible drawback is that it may not be easily generalized to the other methods owing to its dependence on the type of conditional independence test or score function used. For instance, it may not be straightforward to extend this approach to the nonparametric test \citep{Zhang2012kernel} or nonparametric score functions \citep{Huang2018generalized}. Second, some of the secure multiparty computation methods, such as the Paillier's scheme \citep{Paillier1999public} for additively homomorphic encryption, are computationally expensive as it relies on a number of modular multiplications and exponentiations with a large exponent and modulus \citep{Zhang2020batchcrypt}.

\section{Distributed BNSL with ADMM}\label{sec:federated_bnsl}
As described in Section \ref{sec:suff_stats_encryption}, there are several drawbacks for the approach that relies on secure computation to compute the sufficient statistics. Therefore, as is typical in the federated learning setting, in Section \ref{sec:admm} we develop a distributed approach that consists of training local models and exchanging model parameters, with a focus on the linear case for simplicity. We then describe its extension to the other cases in Section \ref{sec:admm_nonlinear}.

\subsection{Linear Case}\label{sec:admm}
To perform federated learning on the structure of BN, BNSL method based on continuous optimization is a natural ingredient as most of the federated learning approaches developed are based on continuous optimization (see \citep{Yang2019federated,Li2020federated,Kairouz2021advances} for a review). In particular, our approach is based the NOTEARS method proposed by \citet{Zheng2018notears} that formulates structure learning of linear BNs as a continuous constrained optimization problem
\begin{equation}
\begin{aligned}
\label{eq:notears}
\min_{B} \quad & \ell(B; \mathbf{x}) + \lambda \|B\|_1  \\
\text{subject~to} \quad & h(B) \coloneqq \operatorname{tr}\left(e^{{B}\odot{B}}\right) - d = 0,
\end{aligned}
\end{equation}
where 
\begin{equation}\label{eq:least_squares}
\ell(B; \mathbf{x}) = \sum_{k=1}^{K}\ell(B; \mathbf{x}_k) = \frac{1}{2n}\sum_{k=1}^{K}\sum_{i=1}^{n_k}\|x_{k,i} - B^\T x_{k,i} \|_2^2
\end{equation}
is the least square loss, $\lambda$ denotes the $\ell_1$ regularization coefficient, $\|\cdot\|_1$ refers to the $\ell_1$ norm defined element-wise, $\odot$ denotes the Hadamard product, and $h(B)\geq 0$ is the acyclicity term which equals zero if and only if $B$ corresponds to a DAG \citep{Zheng2018notears}. Note that $\ell_1$ penalty has been shown to work well with various continuous optimization methods for BNSL \citep{Zheng2018notears,Zheng2020learning,Ng2020role}, as opposed to the $\ell_0$ penalty used in discrete score-based methods \citep{Chickering2002optimal,Van2013ell_0}.

The formulation \eqref{eq:notears} is not applicable in the federated setting as each client must not disclose its local dataset $\mathbf{x}_k$. To allow the clients to collaborate in learning the BN structure without disclosing their local datasets, several federated or distributed optimization approaches can be used, such as federated averaging \citep{McMahan2017communication} and its generalized version \citep{Reddi2021adaptive}. In this work, we adopt a distributed optimization method known as the ADMM \citep{Glowinski1975admm,Gabay1976dual,Boyd2011admm}, such that only model parameters have to be exchanged during the optimization process. In particular, ADMM is an optimization algorithm that splits the problem into different subproblems, each of which is easier to solve, and has been widely adopted in different areas, such as consensus optimization \citep{Bertsekas1989parallel}.

Using ADMM, we split the constrained problem \eqref{eq:notears} into several subproblems, and employ an iterative message passing procedure to obtain to a final solution. ADMM can be particularly effective when a closed-form solution exists for the subproblems, which we are able to derive for the first one. To formulate problem \eqref{eq:notears} into an ADMM form, the problem can be written with local variables $B_1,\dots,B_K\in\mathbb{R}^{d\times d}$ and a common global variable $\W\in\mathbb{R}^{d\times d}$ as its equivalent problem
\begin{equation}
\begin{aligned}\label{eq:notears_admm}
\min_{B_1, \dots, B_K, \W} \quad & \sum_{k=1}^{K}\ell(B_k; \mathbf{x}_k) + \lambda \|\W\|_1\\
\text{subject~to} \quad & h(\W) = 0, \\
	                        & B_k = \W, \, \quad k = 1, \dots, K.
\end{aligned}
\end{equation}
The local variables $B_1,\dots,B_K$ correspond to the model parameters of different clients. Notice that the above problem is similar to the global variable consensus ADMM described by \citet{Boyd2011admm} with an additional constraint $h(\W)=0$ that enforces the global variable $\W$ to represent a DAG. The constraints $B_k=\W,k=1,\dots,K$ are used to ensure that the local model parameters of different clients are equal.

As is typical in the ADMM setting, and similar to NOTEARS, we adopt the augmented Lagrangian method to solve the above constrained minimization problem. It is a class of optimization algorithm that converts the constrained problem into a sequence of unconstrained problems, of which the solutions, under certain conditions, converge to a stationary point of the constrained problem \citep{Bertsekas1982constrained,Bertsekas1999nonlinear}. In particular, the augmented Lagrangian is given by
\begin{flalign*}
&L(B_1, \dots, B_K, \W, \alpha, \beta_1,\dots,\beta_K; \rho_1,\rho_2)\nonumber\\
&= \sum_{i=1}^{K}\ell(B_k; \mathbf{x}_k) + \lambda \|\W\|_1 + \alpha h(\W) + \frac{\rho_1}{2}h(\W)^2 \nonumber\\
&\qquad+ \sum_{k=1}^{K} \Tr\left(\beta_k (B_k - \W)^\T\right) + \frac{\rho_2}{2}\sum_{k=1}^{K}\|B_k - \W\|_F^2,\nonumber
\end{flalign*}
where $\rho_1,\rho_2>0$ are the penalty coefficients, and $\alpha\in\mathbb{R}$ and $\beta_1,\dots,\beta_K\in\mathbb{R}^{d\times d}$ are the estimations of the Lagrange multipliers. We then obtain the following iterative update rules of ADMM:
\begin{flalign}
B_k^{t+1} & \coloneqq \argmin_{B_k} \Big( \ell(B_k; \mathbf{x}_k) + \Tr\left(\beta_k^t (B_k - \W^t)^\T\right)\nonumber\\
&\qquad\qquad\qquad + \frac{\rho_2^t}{2}\|B_k - \W^t\|_F^2 \Big), \label{eq:admm_B_step}\\
\W^{t+1} & \coloneqq \argmin_{W} \bigg( \lambda \|\W\|_1 + \alpha^t h(\W) + \frac{\rho_1^t}{2}h(\W)^2 \nonumber\\
&\qquad\qquad\qquad +\sum_{k=1}^{K} \Tr\left(\beta_k^t (B_k^{t+1} - \W)^\T\right) \nonumber\\
&\qquad\qquad\qquad + \frac{\rho_2^t}{2}\sum_{k=1}^{K}\|B_k^{t+1} - \W\|_F^2 \bigg),\label{eq:admm_M_step}\\
\alpha^{t+1} & \coloneqq \alpha^t + \rho_1^t h\left(\W^{t+1}
\right),\label{eq:admm_alpha_step}\\
\beta_k^{t+1} & \coloneqq \beta_k^t + \rho_2^t \left(B_k^{t+1}-\W^{t+1}\right), \label{eq:admm_beta_step}\\
\rho_1^{t+1} & \coloneqq \gamma_1 \rho_1^t,\label{eq:admm_rho_1_step}\\
\rho_2^{t+1} & \coloneqq \gamma_2 \rho_2^t,\label{eq:admm_rho_2_step}
\end{flalign}
where $\gamma_1,\gamma_2\in\mathbb{R}$ are hyperparameters that control how fast the coefficients $\rho_1,\rho_2$ are increased, respectively.

As described, ADMM is especially effective if a closed-form solution exists for the above optimization subproblems. Notice that the subproblem \eqref{eq:admm_B_step} corresponds to a \emph{proximal minimization} problem and is well studied in the literature of numerical optimization \citep{Combettes2011proximal,Parikh2014proximal}. Let $S_k =(1/n) \sum_{i=1}^{n_k}x_{k,i}x_{k,i}^\T$ and since $S_k + \rho_2^t I$ is invertible, the closed-form solution of problem \eqref{eq:admm_B_step} is
\begin{equation}\label{eq:admm_B_solution}
B_k^{t+1} \coloneqq (S_k + \rho_2^t I)^{-1} (\rho_2^t \W^t - \beta_k^t +S_k),
\end{equation}
with a derivation given in Appendix \ref{sec:derivation_admm_solution}.

Due to the acyclicity term $h(\W)$, we are not able to derive a closed-form solution for problem \eqref{eq:admm_M_step}. One could instead use first-order (e.g., gradient descent) or second-order (e.g., L-BFGS \citep{Byrd2003lbfgs}) method to solve the optimization problem. Here, we follow \citet{Zheng2018notears} and use the L-BFGS method. To handle the $\ell_1$ penalty term, we use the subgradient method \citep{Boyd2003subgradient} to simplify our procedure, instead of the bound-constrained formulation adopted by \citet{Zheng2018notears,Zheng2020learning}. Furthermore, since the acyclicity term $h(\W)$ proposed by \citet{Zheng2018notears} is nonconvex, problem \eqref{eq:admm_M_step} can only be solved up to a stationary solution, which, however, are shown to lead to a good empirical performance in Section \ref{sec:exp}.

\begin{algorithm}[t]
\caption{Distributed BNSL with ADMM}
\label{alg:admm_bnsl}
\begin{algorithmic}[1]
    \Require initial parameters $\rho_1^1,\rho_2^1,\alpha^1,\beta_1^1,\dots,\beta_K^1$; multiplicative factors $\gamma_1,\gamma_2 > 1$; initial point $\W^1$
    \For{$t=1,2,\ldots$}
        \State Each client solves problem \eqref{eq:admm_B_step} in parallel
        \State {\spaceskip  0.67em \relax Central server collects $B_1^{t+1},\dots,B_K^{t+1}$ from}
        \Statex \qquad \quad all clients
        \State Central server solves problem \eqref{eq:admm_M_step}
        \State Central server sends $\W^{t+1}$ to all clients
        \State {\spaceskip  0.7em \relax Central server updates ADMM paramaters}
        \Statex \qquad \quad according to Eqs. \eqref{eq:admm_alpha_step}, \eqref{eq:admm_beta_step},  \eqref{eq:admm_rho_1_step}, \eqref{eq:admm_rho_2_step}
        \State {\spaceskip  1.02em \relax Each client updates ADMM paramaters}
        \Statex \qquad \quad according to Eqs. \eqref{eq:admm_beta_step}, \eqref{eq:admm_rho_2_step}
    \EndFor
\end{algorithmic}
\end{algorithm}

The overall distributed optimization procedure is described in Algorithm \ref{alg:admm_bnsl}, which iterates between client and server updates. Since problem \eqref{eq:admm_B_step} and its closed-form solution \eqref{eq:admm_B_solution} involve only the local dataset of a single client, each client computes its own solution $B_k^{t+1}$ in each iteration, and send it to the central server. With the updated local parameters $B_1^{t+1},\dots,B_K^{t+1}$ from the clients, the central server solves problem \eqref{eq:admm_M_step} for the updated global parameter $\W^{t+1}$, and send it back to each of the clients. The clients and server then update the Lagrange multipliers and the penalty coefficients, and enter the next iteration. The optimization process proceeds in such a way that the local parameters $B_1^t,\dots,B_k^t$ and global parameter $\W^t$ converge to (approximately) the same value, and that $\W^t$ converges to (approximately) a DAG, which is the final solution of the ADMM problem. In this way, each client discloses only its model parameter $B_k^t$, but not the other information or statistics related to its local dataset.

{\bf Information exchange.} \ \ \ 
Within the proposed framework, each client has to transfer the local model parameters, i.e., linear coefficients in the linear case or multilayer perceptron (MLP) parameters in the nonlinear case (see Section \ref{sec:admm_nonlinear}), to the central server in each round of the optimization process, as described in Algorithm \ref{alg:admm_bnsl}. Loosely speaking, the information of the local parameters is exchanged through the global parameter $W$. For the existing distributed BNSL methods (e.g., voting), each client learns a DAG on its own and exchanges only the final estimated DAG.

{\bf Bias terms.} \ \ \ 
\citet{Zheng2018notears} adopt a pre-processing step that centers the data $\mathbf{x}$ before solving problem \eqref{eq:notears}, which is equivalent to adding a bias term to the linear regression involved in the least squares \eqref{eq:least_squares}. To take it into account in the distributed setting, one may similarly add the local bias terms $b_1,\dots,b_K\in\mathbb{R}^d$ to the ADMM problem \eqref{eq:notears_admm}, each of which is owned by a client, along with the additional constraints $b_k=w,k=1,\dots,K$, where $w\in\mathbb{R}^d$ is the global variable corresponding to the bias terms. Similar iterative update rules for $b_1,\dots,b_K,w$ can then be derived. Note however that the bias terms may reveal information related to the variable means of each client's local dataset. Alternatively, each client may center its local dataset $\mathbf{x}_k$ before collectively solving the problem \eqref{eq:notears_admm}, which we adopt in our implementation. 

\subsection{Extension to Other Cases}\label{sec:admm_nonlinear}
We describe how to extend the proposed distributed BNSL approach to the other cases to demonstrate its flexibility. First notice that the update rule \eqref{eq:admm_B_step} is the key step of ADMM to enable distributed learning across different clients as its optimization subproblem involves only the local model parameter $B_k$ and the local dataset $\mathbf{x}_k$ of a single client. Thus, as long as the objective function $\ell(B; \mathbf{x})$ is \emph{decomposable} w.r.t. different clients or different samples, e.g., the least squares loss, we are able to derive similar update rule from the augmented Lagrangian $L(B_1, \dots, B_K, \W, \alpha, \beta_1,\dots,\beta_K; \rho_1,\rho_2)$ w.r.t. the local model parameter $B_k$, and the proposed ADMM approach, given a proper acyclicity constraint term $h(\W)$, will be applicable.

Specifically, minimizing the least squares loss $\ell(B; \mathbf{x})$ is equivalent to linearly regressing each variable on the other variables, which, with the $\ell_1$ penalty, can be considered as a way to estimate the Markov blanket of each variable \citep{Meinshausen2006high,Schmidt2007learning}. Further under the acyclicity constraint, \citet{Aragam2019globally} established the high-dimensional consistency of the global optimum of the $\ell_1$-penalized least squares for learning linear Gaussian BNs with equal noise variances. As suggested by \citet{Zheng2020learning}, one could replace the linear regression with the other suitable model family, as long as it is differentiable and supports gradient-based optimization. In this case, the local variables $B_1,\dots,B_K$ correspond to the parameters of the chosen model family, while the dimension of the global variable $\W$ is equal to that of $B_k$. Similar to problem \eqref{eq:notears_admm}, a constraint function $h$ is required to enforce acyclicity for variable $\W$, and can be constructed using the general procedure described by \citet{Zheng2020learning}. With a proper model family and constraint function $h$, the rest of the ADMM procedure can be directly applied, except that there may not always be a closed-form solution for problem \eqref{eq:admm_B_step}, when, e.g., applied to the nonlinear case involving MLPs.

{\bf Nonlinear case.} \ \ \ 
One can adopt the MLPs for modeling nonlinear dependencies, similar to \citet{Yu19daggnn,Zheng2020learning,Lachapelle2020grandag,Ng2022masked}. Their proposed DAG constraints can also be applied here w.r.t. the global variable $\W$. In particular, following \citet{Zheng2020learning,Lachapelle2020grandag}, we construct an equivalent adjacency matrix from the parameters of MLPs, and formulate the constraint function $h$ based on it. Following \citet{Ng2022masked}, one could also use an additional (approximately) binary matrix that represents the adjacency information with Gumbel-Softmax \citep{Jang2017gumbelsoftmax}, from which the constraint function $h$ is constructed.

{\bf Time-series data.} \ \ \ 
The proposed approach can also be extended to the setting in which each client has a number of independent realizations of time-series data over $d$ variables, and intend to collectively learn a dynamic BN consisting of instantaneous and time-lagged dependencies. In particular, \citet{Pamfil2020dynotears} proposed a method that estimates the structure of a structural vector autoregressive model by minimizing the least squares loss subject to an acyclicity constraint defined w.r.t. the instantaneous coefficients. Since the objective function is decomposable w.r.t. different samples, our ADMM approach is applicable here, in which the global variable $\W$ consists of the instantaneous and time-lagged coefficients, and the constraint function $h$ is defined w.r.t. the part of variable $\W$ that corresponds to the instantaneous coefficients.
\begin{figure*}[!t]
\centering
\includegraphics[width=0.91\textwidth]{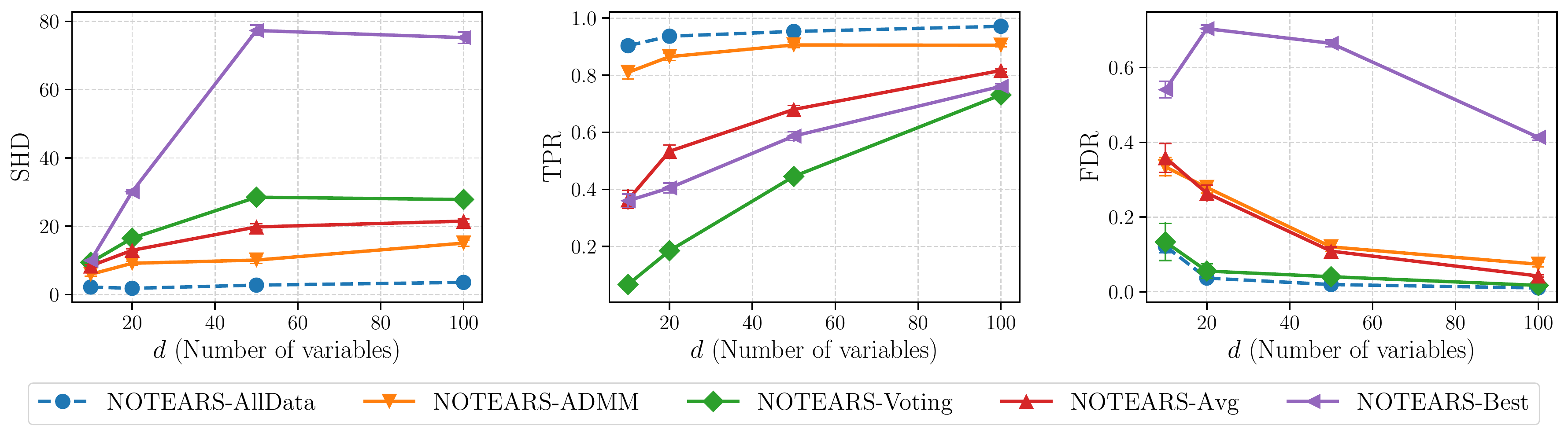}
\caption{Structure learning results of linear Gaussian BNs with varying number of variables. There are $n=3d$ samples in total, distributed evenly across $K=10$ clients. Error bars refer to the standard errors computed over $30$ random runs.}
\label{fig:gaussian_different_nodes}
\end{figure*}

\section{Experiments}\label{sec:exp}
We conduct empirical studies to verify the effectiveness of the proposed federated BNSL approach. In Sections \ref{sec:exp_gaussian_different_nodes} and \ref{sec:exp_gaussian_different_clients}, we provide experimental results with varying number of variables and clients, respectively. We then apply the proposed approach to real data in Section \ref{sec:exp_sachs_different_clients}, and to the nonlinear case in Section \ref{sec:exp_mlp_different_clients}.

{\bf Simulations.} \ \ \ 
The true DAGs are simulated using the Erd\"{o}s--R\'{e}nyi \citep{Erdos1959random} model with number of edges equal to the number of variables $d$. In the linear case, we follow \citet{Zheng2018notears} and generate the data according to the linear Gaussian BNs, in which the coefficients are sampled uniformly at random from $[-2, -0.5] \cup [0.5, 2]$, and the noise terms are standard Gaussian. In the nonlinear case, similar to \citep{Zheng2020learning}, we simulate the nonlinear BNs with each function being a randomly initialized MLP consisting of one hidden layer and $100$ sigmoid activation units. The structures of these two data models are fully identifiable \citep{Peters2013identifiability, Peters2014causal}. We focus on the setting in which each client has a limited sample size, which may in practice be the reason why the clients intend to collaborate.

{\bf Methods.} \ \ \ 
In the linear case, we compare the ADMM approach described in Section \ref{sec:admm}, denoted as NOTEARS-ADMM, to the voting approach proposed by \citet{Na2010distributed}, denoted as NOTEARS-Voting. In particular, we apply the NOTEARS \citep{Zheng2018notears} method to estimate the DAG from each client's local dataset independently, and use a voting scheme that picks those edges identified by more than half of the clients. We also include another baseline that computes the average of the weighted adjacency matrices estimated by NOTEARS from the local datasets, and then perform thresholding to identify the edges, which we denote by NOTEARS-Avg. Apart from voting and averaging, we adopt another baseline that picks the best graph, i.e., with the lowest structural Hamming distance (SHD), from the independently estimated graphs by the clients, denoted as NOTEARS-Best. In practice, the ground truth may not be known, and we are not able to pick the best graph in such a way; however, this serves as a reasonable baseline to our approach. Also, note that the final graph returned by NOTEARS-Voting and NOTEARS-Avg may contain cycles, but we do not apply any post-processing step to remove them as it may worsen the performance. We also report the performance of NOTEARS on all data by combining all clients' local datasets, denoted by NOTEARS-AllData. Similarly, in the nonlinear case, we compare among NOTEARS-MLP-ADMM, NOTEARS-MLP-Voting, NOTEARS-MLP-Avg, NOTEARS-MLP-Best, and NOTEARS-MLP-AllData.

Following \citet{Zheng2018notears,Zheng2020learning}, a thresholding step at $0.3$ is used to remove the estimated edges with small weights. Further implementation details and hyperparameters of the proposed approach are described in Appendix \ref{sec:supp_implementation_details}. The code is available at \url{https://github.com/ignavierng/notears-admm}.

{\bf Metrics.} \ \ \
We use the SHD, true positive rate (TPR), and false discovery rate (FDR) to evaluate the estimated graphs, computed over $30$ random runs.

\subsection{Varying Number of Variables}\label{sec:exp_gaussian_different_nodes}
We consider the linear Gaussian BNs with a total number of $n=3d$ samples distributed evenly across $K=10$ clients, i.e., each client has $\left \lfloor 0.3d\right \rfloor$ samples. We conduct experiments with $d\in\{10,20,50,100\}$ variables.

The results are shown in Figure \ref{fig:gaussian_different_nodes}. One observes that NOTEARS-ADMM performs the best across different number of variables, and the difference of SHD grows as the number of variables increases. It also has relatively high TPRs which are close to those of NOTEARS-AllData, compared to the other baselines, indicating that it manages to identify most of the edges. Among the baselines, NOTEARS-Avg has lower SHDs than NOTEARS-Voting as the latter identifies only a small number of edges, leading to low TPRs. Notice that NOTEARS-Best has high SHDs owing to high FDRs. A possible reason is that the sample size is too small which gives rise to a high estimation error.

{\bf Other baselines.} \ \ \
We also consider BNSL methods that are not based on continuous
optimization. In particular, we apply the voting scheme to the structures independently estimated by PC \citep{Spirtes1991pc} and FGES \citep{Ramsey2017million} from the clients' local datasets, denoted as PC-Voting and FGES-Voting. Since these methods output the Markov equivalence classes instead of DAGs, we follow \citet{Zheng2018notears} and consider each undirected edge as a true positive if there is a directed edge in place of the undirected one in the true DAG. The results are reported in Figure \ref{fig:gaussian_different_nodes_non_continuous} in Appendix \ref{sec:supp_exp_results}, showing that these methods have worse SHDs and TPRs than those of NOTEARS-ADMM.

\begin{figure*}[!t]
\centering
\includegraphics[width=0.91\textwidth]{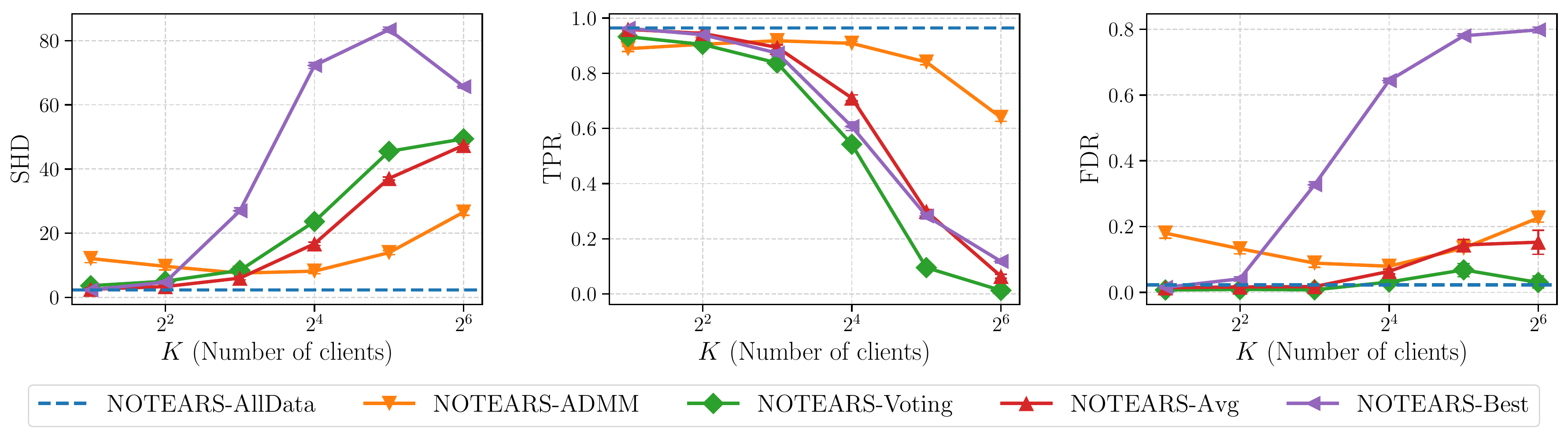}
\caption{Structure learning results of linear Gaussian BNs with $d=50$ variables and varying number of clients. There are $n=256$ samples in total, distributed evenly across $K\in\{2, 4, 8, 16, 32, 64\}$ clients. Error bars refer to the standard errors computed over $30$ random runs.}
\label{fig:gaussian_different_clients_50nodes}
\end{figure*}

\subsection{Varying Number of Clients}\label{sec:exp_gaussian_different_clients}
We now consider a fixed total number of samples which are distributed across different number of clients. For $d\in\{10,20,50\}$ variables, we generate $n=256$ samples and distribute them across $K\in\{2, 4, 8, 16, 32, 64\}$ clients. This may be a challenging setup because, for, e.g., $K=64$ clients, each client has only $4$ samples.

Figure \ref{fig:gaussian_different_clients_50nodes} shows the results with $50$ variables, while those with $d\in\{10,20\}$ variables are depicted by Figure \ref{fig:gaussian_different_clients} in Appendix \ref{sec:supp_exp_results} owing to space limit. As the number of clients $K$ increases, the TPRs of NOTEARS-Voting, NOTEARS-Avg, and NOTEARS-Best quickly deteriorate, leading to high SHDs. On the other hand, although the TPRs of NOTEARS-ADMM also decrease with a larger $K$, they are still relatively high as compared to those of the other baselines. For instance, with $d=20$ variables and $K=64$ clients, NOTEARS-ADMM achieves a TPR of $0.78$, while NOTEARS-Voting, NOTEARS-Avg, and NOTEARS-Best have TPRs of $0.05$, $0.25$, and $0.33$, respectively. This verifies the effectiveness of NOTEARS-ADMM in the setting with a large number of (up to $64$) clients and indicates that exchanging information during the optimization process is a key to learning an accurate BN structure, as is the case for NOTEARS-ADMM. On the contrary, the other baselines utilize only the information of the independently estimated DAGs by the clients, and therefore the information exchange is rather limited.

Interestingly, when the number of clients $K$ is small, the other baselines perform better than NOTEARS-ADMM. For instance, with $50$ variables, NOTEARS-Voting, NOTEARS-Avg, and NOTEARS-Best have better SHDs than NOTEARS-ADMM when there are only $2$ or $4$ clients. This is not surprising because for a small number of clients $K$, the sample size may be sufficient for each client to independently learn an accurate enough BN structure, i.e., $n=128,64$ samples for $K=2,4$ clients, respectively. On the other hand, NOTEARS-ADMM involves solving a more complex nonconvex optimization problem, and therefore the solutions obtained may not be as accurate as those by the baselines especially when the task is relatively easy, i.e., each client has a sufficient sample size and is able to learn an accurate enough BN structure on their own.

\begin{figure}[!t]
\centering
\includegraphics[width=0.41\textwidth]{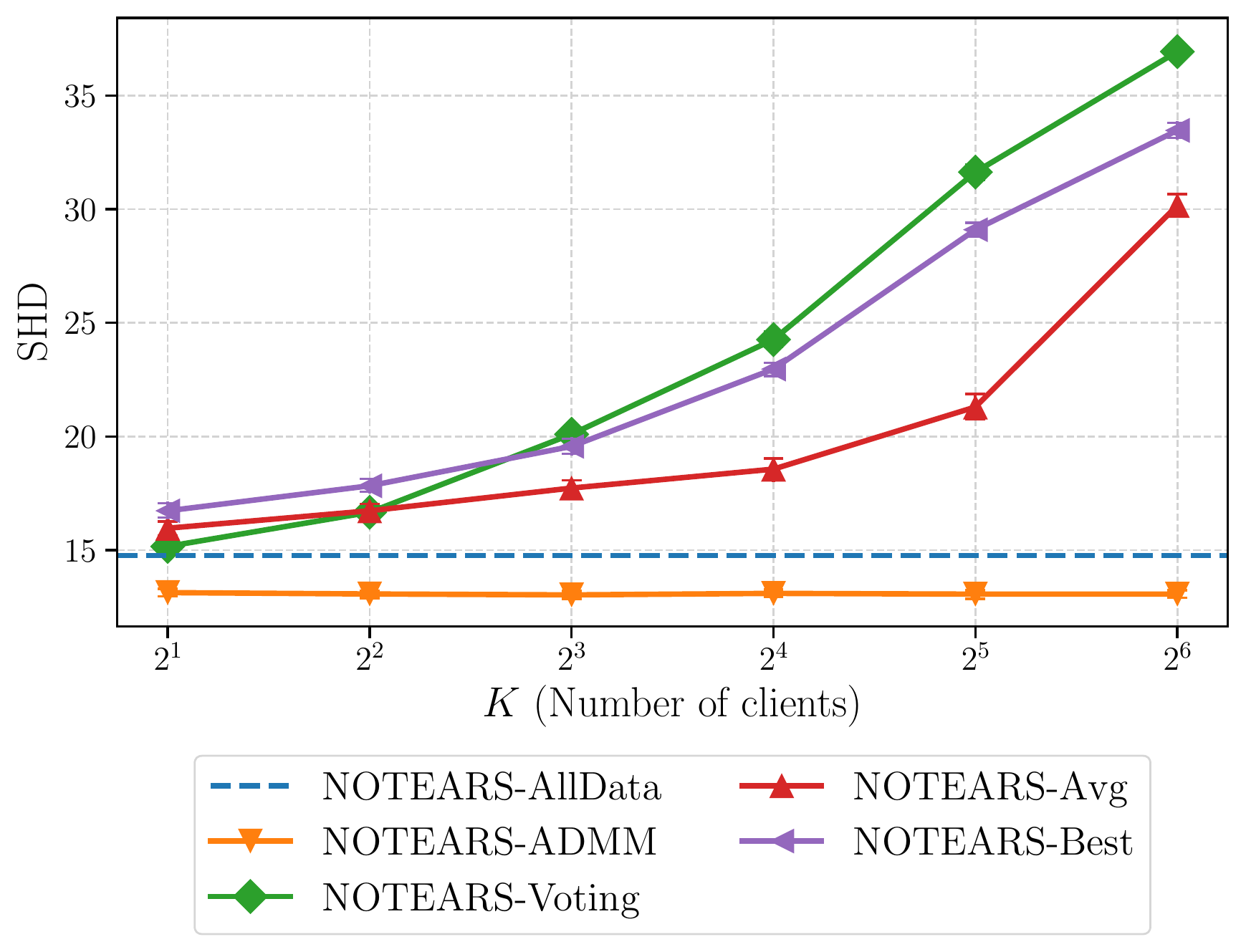}
\caption{Real data with $512$ samples in total, distributed evenly across $K\in\{2, 4,8,16,32, 64\}$ clients.}
\label{fig:sachs_different_clients_shd}
\end{figure}

\subsection{Real Data}\label{sec:exp_sachs_different_clients}
We evaluate the proposed method on a protein expression dataset from \citet{Sachs2005causal}. This dataset contains $n=853$ observational samples and $11$ variables, and the proposed ground truth DAG contains $17$ edges. For each of the $30$ random runs, we randomly pick $n=512$ samples from the dataset and distribute them across $K\in\{2, 4, 8, 16, 32, 64\}$ clients, in order to simulate a setting in which the real data is distributed across different parties.

The SHDs are shown in Figure \ref{fig:sachs_different_clients_shd}, while the complete results including TPRs and FDRs are reported in Figure \ref{fig:sachs_different_clients} in Appendix \ref{sec:supp_exp_results}. Similar to the empirical study in Section \ref{sec:exp_gaussian_different_clients}, we observe here that the performance of NOTEARS-Voting, NOTEARS-Avg, and NOTEARS-Best degrades much as the number of clients increases, whereas NOTEARS-ADMM has an improved and stable performance across different number of clients. This indicates that NOTEARS-ADMM is relatively robust in practice even when there is a large number of clients and each has a small sample size. It is interesting to observe that NOTEARS-ADMM has lower SHDs than those of NOTEARS-AllData, possibly because of the model misspecification on this real dataset.

\subsection{Nonlinear Case}\label{sec:exp_mlp_different_clients}
We now provide empirical results in the nonlinear case to demonstrate the flexibility of the proposed approach. Similar to Section \ref{sec:exp_gaussian_different_clients}, we simulate $512$ samples using the nonlinear BNs with MLPs, in which the ground truths have $50$ variables, and distribute these samples across $K\in\{2, 4, 8, 16\}$ clients. We did not include the experiments for $32$ and $64$ clients as the running time of some of the baselines may be too long.

Due to space limit, we report the SHDs in Figure \ref{fig:mlp_different_clients_shd} and the complete results in Figure \ref{fig:mlp_different_clients} in Appendix \ref{sec:supp_exp_results}. Consistent with previous observations, NOTEARS-MLP-ADMM outperforms the other baselines and has a much stable performance across different number of clients, which is close to that of NOTEARS-MLP-AllData. This demonstrates the effectiveness of our approach in the nonlinear case, and validates the importance of information exchange during the optimization. 
\begin{figure}[!h]
\centering
\includegraphics[width=0.41\textwidth]{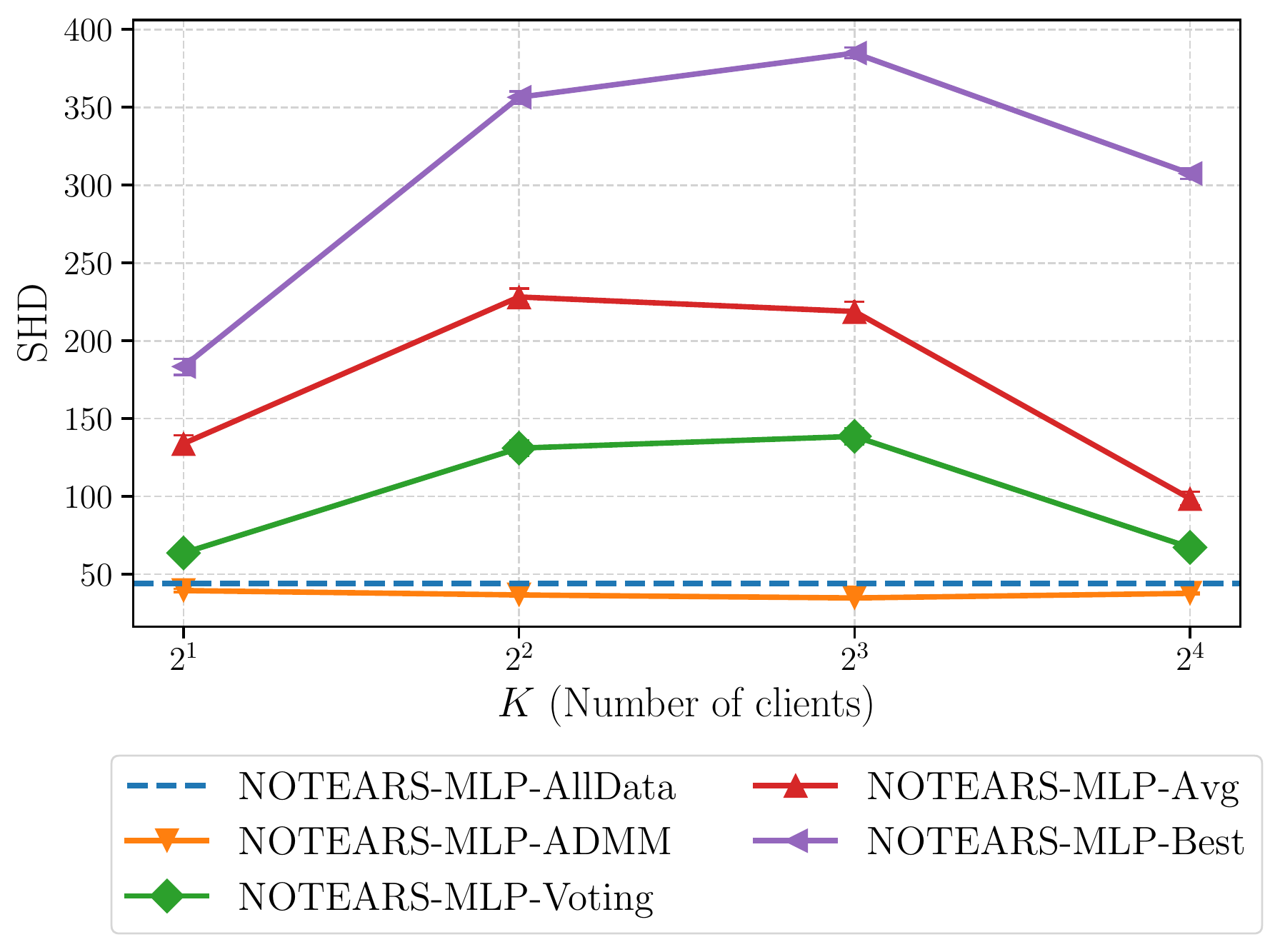}
\caption{Nonlinear BNs with $512$ samples in total, distributed evenly across $K\in\{2, 4,8, 16\}$ clients.}
\label{fig:mlp_different_clients_shd}
\end{figure}
\section{Discussion}\label{sec:discussion}
We presented a federated learning approach to carry out BNSL from horizontally partitioned data. In particular, we developed a distributed BNSL method based on ADMM such that only the model parameters have to be exchanged during the optimization process. The proposed approach is flexible and can be adopted for both linear and nonlinear cases. Our experiments show that it achieves an improved performance over the other methods, especially when there is a relatively large number of clients and each has a small sample size, which is typical in the federated learning setting and may in practice be the reason why the clients intend to collaborate. We discuss below some limitations of the proposed approach and possible directions for future work. We hope that this work could spur future studies on developing federated approaches for BNSL.

{\bf More complex settings.} \ \ \ 
As described in Section \ref{sec:related_work_continuous_optimization}, continuous optimization methods for BNSL have been extended to different settings, such as those with confounders \citep{Bhattacharya2020differentiable} and interventional data \citep{brouillard2020differentiable}. Furthermore, the data distribution of different clients may, in practice, be heterogenous. An important direction is to extend the proposed approach to these settings.

{\bf Federated optimization.} \ \ \ 
The proposed approach with ADMM is stateful and requires each client to participate in each round of communication, and therefore is only suitable for the cross-silo setting, in which the clients usually refer to different organizations or companies. An important direction for future is to explore the use of other federated optimization techniques such as federated averaging \citep{McMahan2017communication} that allow for stateless clients and cross-device learning.

{\bf Convergence properties.} \ \ \ 
Some recent works have studied the convergence of ADMM in the nonconvex setting \citep{Hong2014convergence,Wang2019global}, and of the continuous constrained formulation of BNSL \citep{Wei2020nofears,Ng2022convergence}. The latter showed that the regularity condition required by standard convergence results of augmented Lagrangian method is not satisfied, and thus its behavior is similar to that of the quadratic penalty method \citep{Powell1969nonlinear, Fletcher1987practical}, another class of algorithm for solving constrained optimization problem. It is an interesting future direction to investigate whether similar convergence properties hold for our approach with ADMM.

{\bf Vertically partitioned data.} \ \ \ 
In practice, the data may be vertically partitioned across different clients, i.e., they may own different variables and wish to collectively carry out BNSL. Therefore, it can be useful to develop a federated approach for the vertical setting. A related setting is to estimate structures from overlapping variables; see a review in Section \ref{sec:related_work_overlapping_variables}.

{\bf Privacy protection.} \ \ \ 
The proposed ADMM procedure involves sharing the model parameters with the central server. Therefore, another important direction is to investigate how much information the model parameters may leak, because they have been shown to possibly leak some information in certain cases, e.g., image data \citep{Phong2018privacy}. It is also interesting to consider the use of differential privacy \citep{Dwork2014algorithmic} or homomorphic encryption, e.g., the Paillier's scheme \citep{Paillier1999public}, for further privacy protection of the model parameters.
\section*{Acknowledgments}
The authors would like to thank the anonymous reviewers for helpful comments and suggestions. This work was supported in part by the National Institutes of Health (NIH) under Contract R01HL159805, by the NSF-Convergence Accelerator Track-D award \#2134901, by the United States Air Force under Contract No. FA8650-17-C7715, and by a grant from Apple. The NIH or NSF is not responsible for the views reported in this article.

\bibliographystyle{abbrvnat}
\bibliography{ms}


\clearpage
\appendix

\thispagestyle{empty}
\onecolumn \makesupplementtitle
\section{Derivation of Closed-Form Solution}\label{sec:derivation_admm_solution}
For completeness, in this section we derive the closed-form solution of the subproblem \eqref{eq:admm_B_step} for our ADMM approach in the linear case. We drop the superscript $t$ to lighten the notation, which leads to the optimization problem
\[
\min_{B_k}\ \ f(B_k)\coloneqq\ell(B_k; \mathbf{x}_k) + \Tr\left(\beta_k (B_k - \W)^\T\right) + \frac{\rho_2}{2}\|B_k - \W\|_F^2.
\]
This is known as a \emph{proximal minimization} problem and is well studied in the literature of numerical optimization \citep{Combettes2011proximal,Parikh2014proximal}. With a slight abuse of notation, let $\mathbf{x}_k\in\mathbb{R}^{n_k\times d}$ denote the design matrix that corresponds to the samples of the $k$-th client. Also let $S_k =(1/n) \mathbf{x}_k^\T \mathbf{x}_k$. The first term of the function $f(B_k)$ can be written as
\begin{flalign*}
\ell(B_k; \mathbf{x}_k) &=\frac{1}{2n}\|\mathbf{x}_k - \mathbf{x}_k B_k\|_F^2\\
&= \frac{1}{2n}\Tr\left((\mathbf{x}_k - \mathbf{x}_k B_k)^\T (\mathbf{x}_k - \mathbf{x}_k B_k)\right)\\
&= \frac{1}{2n}\Tr\left(\mathbf{x}_k^\T \mathbf{x}_k - B_k^\T\mathbf{x}_k^\T\mathbf{x}_k - \mathbf{x}_k^\T \mathbf{x}_k B_k + B_k^\T\mathbf{x}_k^\T\mathbf{x}_k B_k \right)\\
&= \frac{1}{2}\Tr\left(S_k -  B_k^\T S_k - S_k B_k + B_k^\T S_k B_k \right)\\
&= \frac{1}{2}\Tr\left(S_k -  2B_k^\T S_k + B_k^\T S_k B_k \right),
\end{flalign*}
where the last line follows from $S_k$ being symmetric. Similarly, the third term of function $f(B_k)$ can be written as
\begin{flalign*}
\frac{\rho_2}{2}\|B_k - \W\|_F^2 &= \frac{\rho_2}{2}\Tr\left((B_k-W)^\T(B_k-W)\right)\\
&=\frac{\rho_2}{2}\Tr\left(B_k^\T B_k-2B_k^\T W+ W^\T W\right).
\end{flalign*}
Therefore, we have
\begin{flalign*}
f(B_k)&=\frac{1}{2}\Tr\left(S_k -  2B_k^\T S_k + B_k^\T S_k B_k \right) + \Tr\left(\beta_k B_k^\T-\beta_k W^\T\right) + \frac{\rho_2}{2}\Tr\left(B_k^\T B_k-2B_k^\T W+ W^\T W\right)\\
&= \Tr\left(\frac{1}{2}B_k^\T \left(S_k +\rho_2 I\right)B_k-B_k^\T(\rho_2 W-\beta_k+S_k)\right)+\text{const},
\end{flalign*}
and its derivative is given by
\begin{flalign*}
\nabla_{B_k} f(B_k)&= \frac{1}{2}\left(\left(S_k +\rho_2 I\right)^\T + \left(S_k +\rho_2 I\right)\right) B_k -(\rho_2 W-\beta_k +S_k)\\
&= (S_k +\rho_2 I) B_k -(\rho_2 W-\beta_k +S_k).
\end{flalign*}
By definition, we have $\rho_2>0$, which implies that the matrix $S_k +\rho_2 I$ is symmetric positive definite, and therefore is invertible. Solving $\nabla_{B_k} f(B_k)=0$ yields the solution
\[
B_k^* = (S_k +\rho_2 I)^{-1}(\rho_2 W-\beta_k+S_k). 
\]
Since the matrix $(1/2)(S_k +\rho_2 I)$ is symmetric positive definite, the function $f(B_k)$ is strictly convex, indicating that $B_k^*$ is its unique global minimum.

\section{Implementation Details and Hyperparameters}\label{sec:supp_implementation_details}

{\bf Implementation details.} \ \ \ 
The proposed distributed BNSL approach with ADMM is implemented with PyTorch \citep{Paszke2019pytorch}. In both linear and nonlinear cases (corresponding to NOTEARS-ADMM and NOTEARS-MLP-ADMM, respectively), we follow \citet{Zheng2018notears,Zheng2020learning} and use the L-BFGS method \citep{Byrd2003lbfgs} to solve the unconstrained optimization problem in the augmented Lagrangian method, except for problem \eqref{eq:admm_B_step}, for which a closed-form solution exists. To handle the $\ell_1$ penalty term, we use the subgradient method \citep{Boyd2003subgradient} to simplify our procedure, instead of the bound-constrained formulation adopted by \citet{Zheng2018notears,Zheng2020learning}. After the optimization process of ADMM finishes, we use the final solution of the global variable $W$ as our estimated model function, which corresponds to the  linear coefficients and MLP parameters in the linear and nonlinear cases, respectively. For the linear case, the linear coefficients directly represent the weighted adjacency matrix, while for the nonlinear case, similar procedure described by \citet{Zheng2020learning} is used to construct the equivalent weighted adjacency matrix. We then perform thresholding at $0.3$ to obtain the final estimated structure.

{\bf Hyperparameters.} \ \ \ 
We set $\gamma_1$ and $\gamma_2$ to $1.75$ and $1.25$, respectively. The initial estimations of the Lagrange multipliers $\alpha^1$ and $\beta_1^1,\dots,\beta_K^1$ are set to zero and zero matrices, respectively. For the initial penalty coefficients and $\ell_1$ regularization coefficient, we find that $\rho_1^1=\rho_2^1=0.001$ and $\lambda=0.01$ work well in the linear case, while $\rho_1^1=\rho_2^1=0.1$ and $\lambda=0.001$ work well in the nonlinear case. We also set the maximum number of augmented Lagrangian iterations to $200$, and the maximum penalty coefficients to $1\times 10^{16}$. These values are selected using small-scale experiments with $d=30$ variables and $K=8$ clients.

\section{Supplementary Experimental Results}\label{sec:supp_exp_results}
\begin{figure*}[!t]
\centering
\includegraphics[width=1.0\textwidth]{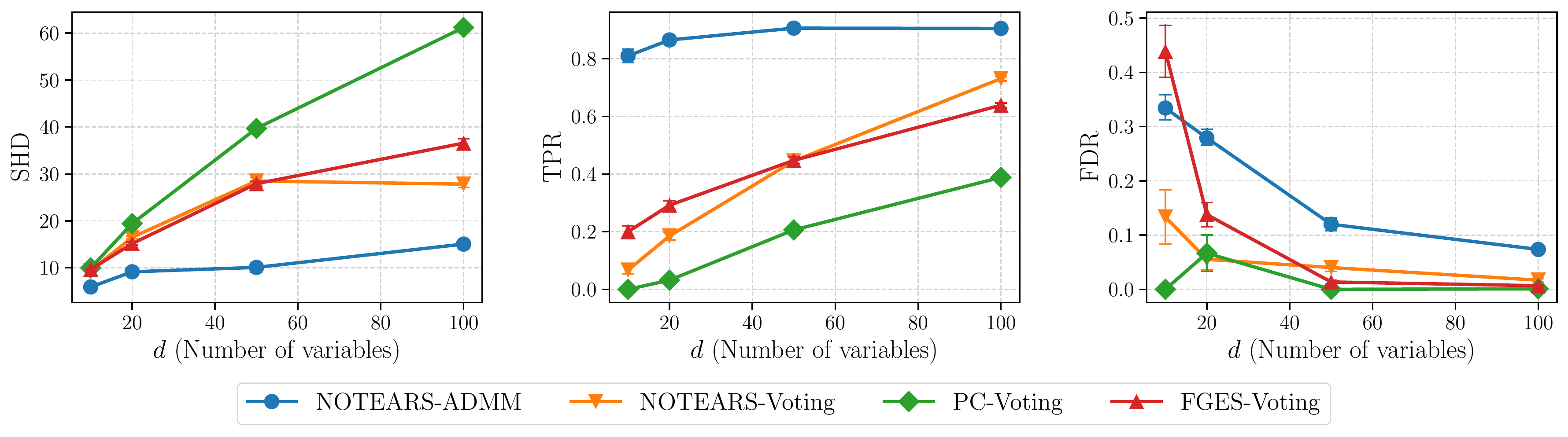}
\caption{Structure learning results of linear Gaussian BNs with varying number of variables. There are $n=3d$ samples in total, distributed evenly across $K=10$ clients. BNSL methods that are not based on continuous optimization are included, i.e., PC-Voting and FGES-Voting. Error bars refer to the standard errors computed over $30$ random runs.}
\label{fig:gaussian_different_nodes_non_continuous}
\end{figure*}

\begin{figure}[!t]
\centering
\subfloat[10 variables.]{
  \includegraphics[width=1.0\textwidth]{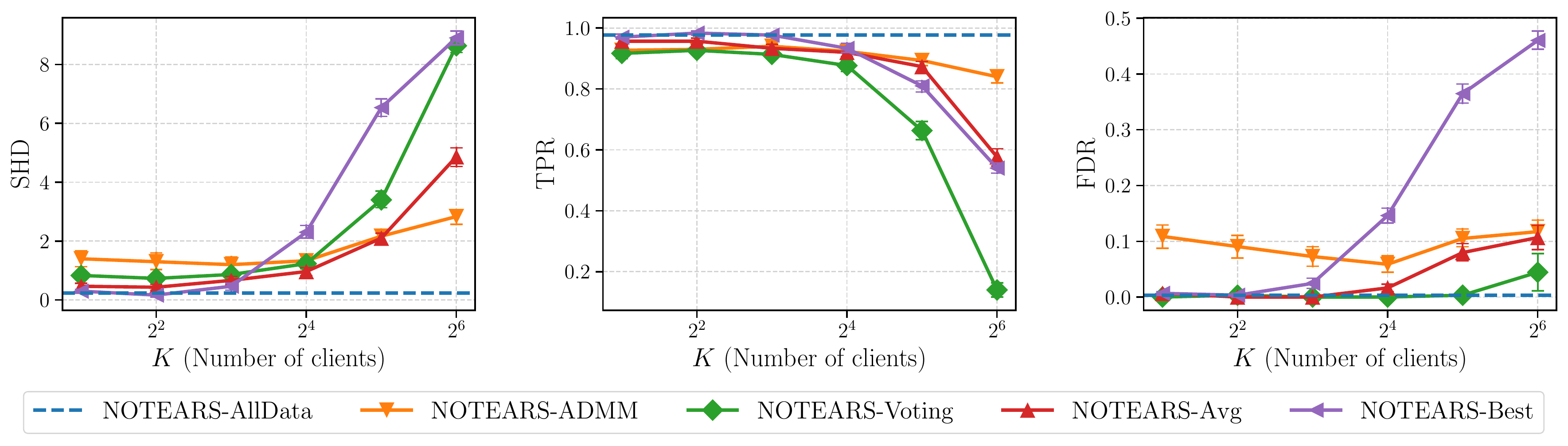}
} \\
\subfloat[20 variables.]{
  \includegraphics[width=1.0\textwidth]{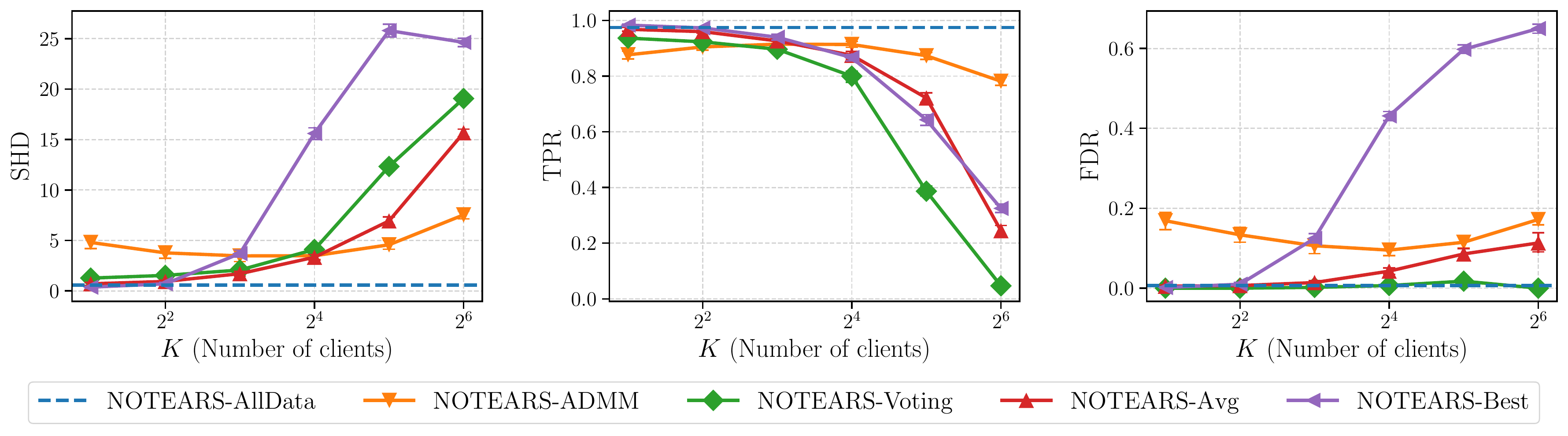}
}
\caption{Structure learning results of linear Gaussian BNs with varying number of clients. There are $n=256$ samples in total, distributed evenly across $K\in\{2, 4, 8, 16, 32, 64\}$ clients. Error bars refer to the standard errors computed over $30$ random runs.}
\label{fig:gaussian_different_clients}
\end{figure}

\begin{figure*}[!t]
\centering
\includegraphics[width=1.0\textwidth]{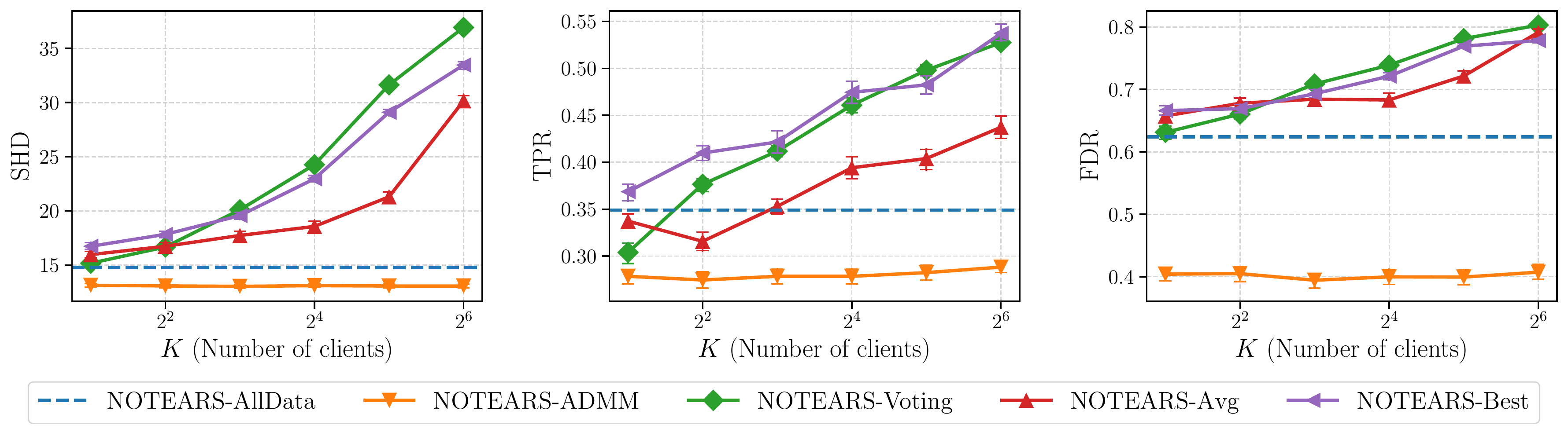}
\caption{Structure learning results of a real dataset with varying number of clients. There are $n=512$ samples in total, distributed evenly across $K\in\{2, 4, 8, 16, 32, 64\}$ clients. Error bars refer to the standard errors computed over $30$ random runs.}
\label{fig:sachs_different_clients}
\end{figure*}

\begin{figure*}[!t]
\centering
\includegraphics[width=1.0\textwidth]{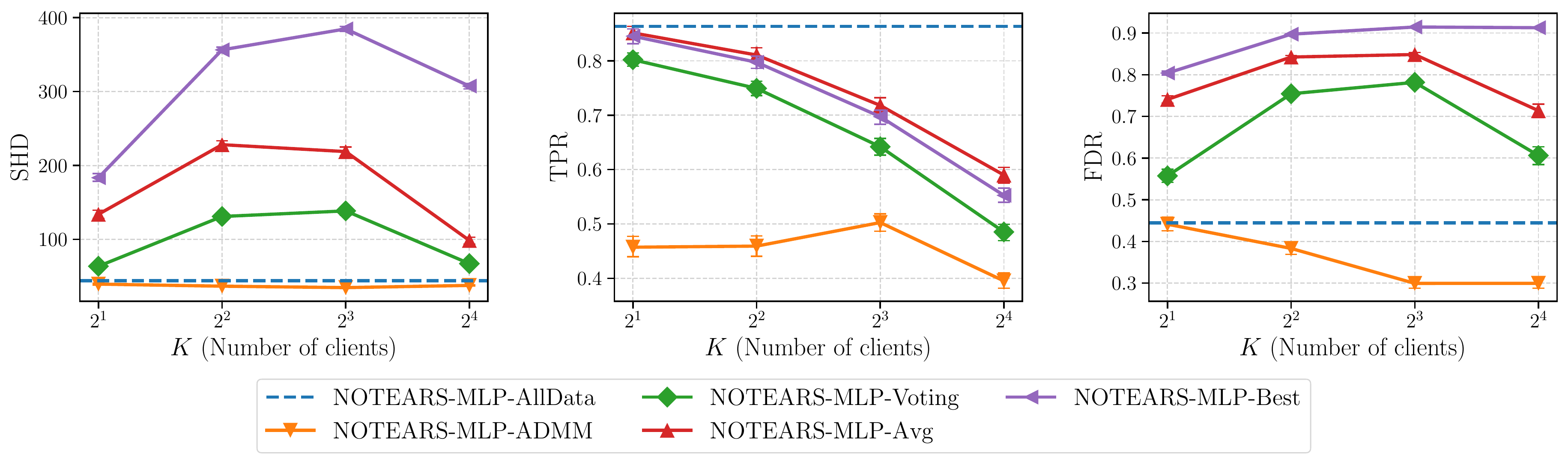}
\caption{Structure learning results of nonlinear BNs with $d=50$ variables and varying number of clients. There are $n=512$ samples in total, distributed evenly across $K\in\{2, 4, 8, 16\}$ clients. Error bars refer to the standard errors computed over $30$ random runs.}
\label{fig:mlp_different_clients}
\end{figure*}

This section provides additional empirical results for Section \ref{sec:exp}, as shown in Figures \ref{fig:gaussian_different_nodes_non_continuous}, \ref{fig:gaussian_different_clients}, \ref{fig:sachs_different_clients}, and \ref{fig:mlp_different_clients}.

\clearpage

\end{document}